\documentclass{article}

\usepackage{microtype}
\usepackage{graphicx}
\usepackage{subfigure}
\usepackage{booktabs} % for professional tables

\usepackage[table]{xcolor}
\definecolor{Gray}{gray}{0.9}
\usepackage{hhline}
\usepackage{multirow, multicol}
\usepackage[toc,titletoc,page]{appendix}

\let\appendixpagenameorig\appendixpagename
\renewcommand{\appendixpagename}{\Large{\appendixpagenameorig}}

\usepackage{times}
\usepackage{epsfig}
\usepackage{graphicx}
\usepackage{amsmath}
\usepackage{amssymb}
\usepackage{mathtools}
\usepackage{enumitem}
\usepackage{balance}

\usepackage{hyperref}

\usepackage[accepted]{arxiv}

\icmltitlerunning{Set Representation Learning with Generalized Sliced Wasserstein Embeddings}

\begin{document}
% \onecolumn

\twocolumn[
\title{\Large \bf Set Representation Learning with Generalized Sliced-Wasserstein Embeddings}
\author{%
{\bf \normalsize Navid Naderializadeh\thanks{\icmlEqualContribution}~, Soheil Kolouri$^*$, Joseph F. Comer$^*$, Reed W. Andrews, \& Heiko Hoffmann}\\
{\normalsize HRL Laboratories, LLC., Malibu, CA 90265}
}
\date{}
\maketitle
% \begin{icmlauthorlist}
% \icmlauthor{Navid Naderializadeh}{equal,to}
% \icmlauthor{Soheil Kolouri}{equal,to}
% \icmlauthor{Joseph F. Comer}{equal,to}
% \icmlauthor{Reed W. Andrews}{to}
% \icmlauthor{Heiko Hoffmann}{to}
% \end{icmlauthorlist}

% \icmlaffiliation{to}{HRL Laboratories, LLC}

\icmlcorrespondingauthor{Soheil Kolouri}{skolouri@hrl.com}
\icmlcorrespondingauthor{Navid Naderializadeh}{nnaderializadeh@hrl.com}

% You may provide any keywords that you
% find helpful for describing your paper; these are used to populate
% the "keywords" metadata in the PDF but will not be shown in the document
\icmlkeywords{Machine Learning, ICML}

\vskip 0.3in
]

\thispagestyle{empty}
% this must go after the closing bracket ] following \twocolumn[ ...

% This command actually creates the footnote in the first column
% listing the affiliations and the copyright notice.
% The command takes one argument, which is text to display at the start of the footnote.
% The \icmlEqualContribution command is standard text for equal contribution.
% Remove it (just {}) if you do not need this facility.

%\printAffiliationsAndNotice{}  % leave blank if no need to mention equal contribution
\printAffiliationsAndNotice{\icmlEqualContribution} % otherwise use the standard text.

%%%%%%%%% ABSTRACT
\begin{abstract}
   
An increasing number of machine learning tasks deal with learning representations from set-structured data. Solutions to these problems involve the composition of permutation-equivariant modules (e.g., self-attention, or individual processing via feed-forward neural networks) and permutation-invariant modules (e.g., global average pooling, or pooling by multi-head attention). In this paper, we propose a geometrically-interpretable framework for learning representations from set-structured data, which is rooted in the optimal mass transportation problem. In particular, we treat elements of a set as samples from a probability measure and propose an exact Euclidean embedding for Generalized Sliced Wasserstein (GSW) distances to learn from set-structured data effectively. We evaluate our proposed framework on multiple supervised and unsupervised set learning tasks and demonstrate its superiority over state-of-the-art set representation learning approaches.

% Wasserstein distances and their variations, like generalized-sliced-Wasserstein (GSW) distances, have received tremendous attention in the machine learning and computer vision communities. In this paper, we treat elements of a set as samples from a probability measure and propose an exact Euclidean embedding for GSW distances to learn from set-structured data effectively.

\end{abstract}

%%%%%%%%% BODY TEXT
\section{Introduction}

Many traditional machine learning architectures, such as feed-forward neural networks, operate on constant-size inputs. Each sample fed into such an architecture consists of a \emph{list} of features, whose size is kept fixed throughout the training/testing process. On the other hand, architectures, such as convolutional and recurrent neural networks (CNNs and RNNs, respectively), allow the size of each input sample to be arbitrary, but they still carry a notion of, e.g., spatial or temporal, ordering among the input features.

Nevertheless, there exist many problems in which each sample consists of an \emph{unordered set} of elements. 3-D point cloud classification, sequence ordering, and even problems as simple as finding the maximum/minimum element of a set are examples of problems in which the size of each input sample can be different and the ordering of the input elements is unimportant. Such a \emph{set learning} phenomenon also arises at the output of graph neural networks (GNNs), as well as CNNs, where a backbone is applied on the graph nodes (resp., original input image pixels), leading to an unordered set of node embeddings (resp., superpixel features). This intermediate set of embeddings are then mapped to a constant-size embedding that represents the entire input graph/image through a pooling method, such as average/max pooling, which is insensitive to the size and ordering of the embedding set.

Such problems have motivated general-purpose \emph{set embedding} methods that provide a parametric mapping of sets to a fixed-dimensional embedding space by means of a \emph{permutation-invariant} function. In~\cite{zaheer2017deep}, the authors introduced the notion of \emph{Deep Sets}, where each element of a set first undergoes a backbone, and the resulting embeddings are then aggregated via a simple mean/sum pooling method. The work in~\cite{lee2019set} proposed \emph{Set Transformers}, where a permutation-equivariant self-attention mechanism is used to perform message-passing among the set elements, followed by a cross-attention module with a set of \emph{seed} elements to derive a permutation-invariant embedding for the whole set. Moreover, in~\cite{skianis2020rep}, a network-flow-based approach is introduced, where the relationship between each set and multiple \emph{hidden sets} are used to derive the set embeddings for any given set.

On a different, but related note, in their pioneering work, \citet{kusner2015word} viewed the elements of a set as samples of an underlying probability measure and leveraged the \emph{1-Wasserstein distance} (i.e., the earth mover's distance) to compare sets with one another. The work of \citet{kusner2015word} and its extension to supervised learning \cite{NIPS2016_10c66082} show that comparing the underlying probability measures is a powerful idea and leads to excellent performance. The Wasserstein distances and their variations have become increasingly popular in machine learning and computer vision \cite{kolouri2017optimal}, e.g., for generative modeling \cite{arjovsky2017wasserstein,gulrajani2017improved,tolstikhin2018wasserstein,kolouri2018sliced} and domain adaptation \cite{courty2017optimal,damodaran2018deepjdot}, among others. These distances have recently been used for measuring distances between graph/image embedding sets ~\cite{zhang2020deepemd,togninalli2019wasserstein, kolouri2021wasserstein}.

% More recently, in~\cite{deep-emd}, Wasserstein distances have been used to treat the inputs/outputs a CNN backbone as sets of patches/embeddings, and in~\cite{kolouri2020wasserstein}, linear Wasserstein embedding is utilized to derive constant-size representations for samples in graph datasets, which are then utilized for downstream graph classification tasks.

The computational complexity of the inherent linear programming involved in calculating the Wasserstein distance has given rise to a large number of works in multiple directions to address this computational challenge, e.g., various convex regularizations \cite{cuturi2013sinkhorn,NIPS2016_2a27b814}, and  multi-scale and hierarchical solvers \cite{oberman2015efficient,schmitzer2016sparse}. Alternatively, the sliced-Wasserstein (SW) distance \cite{deshpande2018generative,Kolouri_2018_CVPR,deshpande2019max}, and generalized sliced-Wasserstein (GSW) distances \cite{kolouri2019generalized} leverage the closed-form solution of the optimal transport problem for one-dimensional distributions to provide a computationally efficient distance that shares some statistical and topological characteristics with the Wasserstein distances \cite{nadjahi2020statistical}.

In this paper, we leverage the GSW distance and propose a geometrically-interpretable framework for learning from set-structured data. We make the following contributions:

% Similar to the linear Wasserstein embedding \cite{wang2013linear,kolouri2017optimal,kolouri2020wasserstein}, which approximates the Wasserstein distance via a logarithmic projection to the tangent space of the  probability measures' manifold at a fixed reference measure, and due to the one-dimensional nature of the slices, the second condition is an equality. 

% Slicing operation(s) have recently gained ample interest and are used in a multitude of applications \cite{}.

% \noindent{\bf Contributions.} our specific contributions in this paper are:
\begin{enumerate}
    \item Proposing a Euclidean embedding for the generalized sliced-Wasserstein distance, which we refer to as Generalized Sliced-Wasserstein Embedding (GSWE);
    \item Leveraging GSWE to devise a new framework for unsupervised and supervised learning on set-structured data; and,
    \item Demonstrating the effectiveness of GSWE in learning from sets and comparing it to state-of-the-art approaches \cite{lee2019set,skianis2020rep}.
\end{enumerate}

\section{Related Work}

\subsection{Pooling Methods for Set Learning}

Permutation-invariant functions are a critical component in learning from sets and are often referred to as pooling layers. Max, min, and sum or average pooling are simple examples of such functions. Recently, various work has shown the effectiveness of more sophisticated and often parametrized pooling operators in improving the performance of learning algorithms \cite{cangea2018towards,murphy2018janossy,lee2019set,Zhang2020FSPool}. Attention-based pooling \cite{trinh2019selfie,lee2019set}, in particular, has been shown to perform really well in practice. In this paper, we introduce a novel pooling using optimal transportation and show that this pooling is geometrically meaningful. More precisely, the proposed process is equivalent to calculating an embedding for sets in which the Euclidean distance is equal to the generalized sliced-Wasserstein distance. Our work is closely related to the concurrent work by \citet{mialon2021a}, however, we arrive at our proposed pooling from a very different perspective compared to \cite{mialon2021a}. In short, \citet{mialon2021a} propose a linear Wasserstein embedding, similar to \cite{kolouri2021wasserstein}, in a reproducing kernel Hilbert space (RKHS), while our proposed framework is based on devising an exact Euclidean embedding for the generalized sliced-Wasserstein distance. In addition, we develop a unique unsupervised learning scheme that is motivated by the concept of optimizing a set of slices, similar to the idea of max-sliced Wasserstein distance \cite{deshpande2019max,kolouri2019generalized}, leveraging the recently developed contrastive learning losses \cite{le2020contrastive}.

\subsection{Self-Supervised Learning}
Learning without or with few labels is the key to unlocking the true potential of deep learning. Self-supervised learning approaches are recently shown to succeed at unsupervised representation learning in many tasks, mainly in computer vision and natural language processing. In this paper, we are interested in self-supervised learning from \emph{set}-structured data.  The essence of self-supervised learning is to utilize a supervision signal that can be programmatically generated from the data without the need for hand-crafted labels.

Many classic self-supervised learning methods employ a so-called proxy- or pseudo-task, which is expected to require the model to learn feature representations that will be useful in the ``downstream'' task, or primary task of interest. In one early example~\cite{larsson2016learning,zhang2016colorful}, it was shown that a model can be pretrained by inducing the network to correctly reproduce the original color in color photographs which have been made black-and-white. Numerous other pseudo-tasks, including rotation prediction (RotNet) \cite{gidaris2018unsupervised}, jigsaw puzzle solving \cite{misra2020self}, and object counting \cite{noroozi2017representation} have been explored and produced promising results.

The crafting of suitable pseudo-tasks for a given dataset and downstream task requires care. For example, RotNet, a pseudo-task which rotates sample images and requires the network to predict which rotation has been applied, has been shown to work very well on vertically-biased natural image data, but would likely produce only a weak effect on rotationally-invariant image data (e.g., aerial images).

Partially in response to this, interest has grown in the so-called contrastive learning methods~\cite{oord2018representation}. The core idea of contrastive learning is to create a latent feature space in which features from similar data are close together,  and features from dissimilar data are spread apart. Notions of “similar” and “dissimilar” vary, but it is common to use augmentations to produce alternative “views” of each data point, and to consider all of the views of a given data point as being like unto one another (positive samples), while views of other data points are dissimilar (negative samples) \cite{le2020contrastive}.

In one early example of contrastive learning, \cite{wu2018unsupervised}, only negative examples were utilized: The latent space was constructed by causing each image’s representation, or latent feature, to lie as far from all other images’ representations as possible within a compact space. Necessarily, visually similar images in the training data begin to clump within the feature space. Later methods, such as MoCo \cite{he2020momentum}, BYOL \cite{grill2020bootstrap}, and SimCLR \cite{chen2020simple} utilize memory banks, momentum decay of network parameters, or very large batch sizes to prevent mode collapse in the feature space due to the use of positive samples. In SimSiam \cite{chen2020exploring}, the authors demonstrate that utilizing a stop-gradient operator in the loss calculation is sufficient to prevent such collapse, and rely on positive samples only. We will show how we can leverage the ideas of SimCLR and SimSiam in our proposed framework for unsupervised representation learning on set-structued data.

%-------------------------------------------------------------------------

\section{Preliminaries}

\subsection{Wasserstein Distances}
Let $\mu_i$ denote a Borel probability measure with finite $p$\textsuperscript{th} moment defined on $\mathcal{Z}\subseteq\mathbb{R}^d$, with corresponding probability density function $q_i$, i.e., $d\mu_i(z)=q_i(z)dz$. The $p$-Wasserstein distance between $\mu_i$ and $\mu_j$ defined on $\mathcal{Z},\mathcal{Z}'\subseteq\mathbb{R}^d$ is the solution to the optimal mass transportation problem with $\ell_p$ transport cost \cite{villani2008optimal}:
\begin{align}
\mathcal{W}_p(\mu_i,\mu_j)=\left(\inf_{\gamma\in \Gamma(\mu_i,\mu_j)} \int_{\mathcal{Z}\times \mathcal{Z}'} \|z-z'\|^p d\gamma(z,z') \right)^{\frac{1}{p}},
\end{align}
where $\Gamma(\mu_i,\mu_j)$ is the set of all transportation plans $\gamma\in\Gamma(\mu_i,\mu_j)$ such that $\gamma(A \times \mathcal{Z}')= \mu_i(A)$ and $\gamma(\mathcal{Z} \times B)= \mu_j(B)$ for any Borel subsets $A\subseteq \mathcal{Z}$ and $B\subseteq \mathcal{Z}'$.
% \begin{alignat}{3}
% \gamma(A \times \mathcal{Z}')&= \mu_i(A) \quad &&\text{for any Borel subset } A\subseteq \mathcal{Z},\\
% \gamma(\mathcal{Z} \times B)&= \mu_j(B)  \quad &&\text{for any Borel subset } B\subseteq \mathcal{Z}'.
% \end{alignat}
Due to Brenier's theorem \cite{brenier1991polar}, for absolutely continuous probability measures $\mu_i$ and $\mu_j$ (with respect to the Lebesgue measure), the $p$-Wasserstein distance can be equivalently obtained from the Monge formulation \cite{villani2008optimal},
\begin{align}
    \mathcal{W}_p(\mu_i,\mu_j)=\left(\operatorname*{inf}_{f\in MP(\mu_i,\mu_j)} \int_{\mathcal{Z}} \|z-f(z)\|^p  d\mu_i(z)\right)^{\frac{1}{p}},
\end{align}
where $MP(\mu_i,\mu_j)=\{ f:\mathcal{Z}\rightarrow \mathcal{Z}' ~|~ f_\#\mu_i=\mu_j\}$ and $f_\#\mu_i$ represents the pushforward of measure $\mu_i$, characterized as $f_\#\mu_i(B)=\mu_i(f^{-1}(B))$ 
for any Borel subset $B\subseteq \mathcal{Z}'$. The mapping $f$ is referred to as a transport map \cite{kolouri2017optimal}, and the optimal transport map is called the Monge map.
% (named after the French mathematician Gaspard Monge).
For discrete probability measures, when the transport plan $\gamma$ is a deterministic optimal coupling, such a transport plan is referred to as a Monge coupling \cite{villani2008optimal}. For one-dimensional probability measures, the Wasserstein distance has a closed-form solution and can be calculated as
\begin{align}
\mathcal{W}_p(\mu_i,\mu_j)=\left(\int_{0}^1|F_{\mu_i}^{-1}(\tau)-F_{\mu_j}^{-1}(\tau)|^pd\tau\right)^{\frac{1}{p}},
\end{align}
where $F_{\mu_i}(t)=\mu_i([-\infty,t])$ and $F_{\mu_i}^{-1}$ is the quantile function of $\mu_i$. The simplicity of calculating Wasserstein distances between one-dimensional probability measures has led to the idea of (max-)sliced \cite{bonnotte2013unidimensional,deshpande2019max} and (max-)generalized-sliced Wasserstein distances \cite{kolouri2019generalized}, which we will review next.  

\begin{figure}[t!]
    \centering
    \includegraphics[width=\columnwidth]{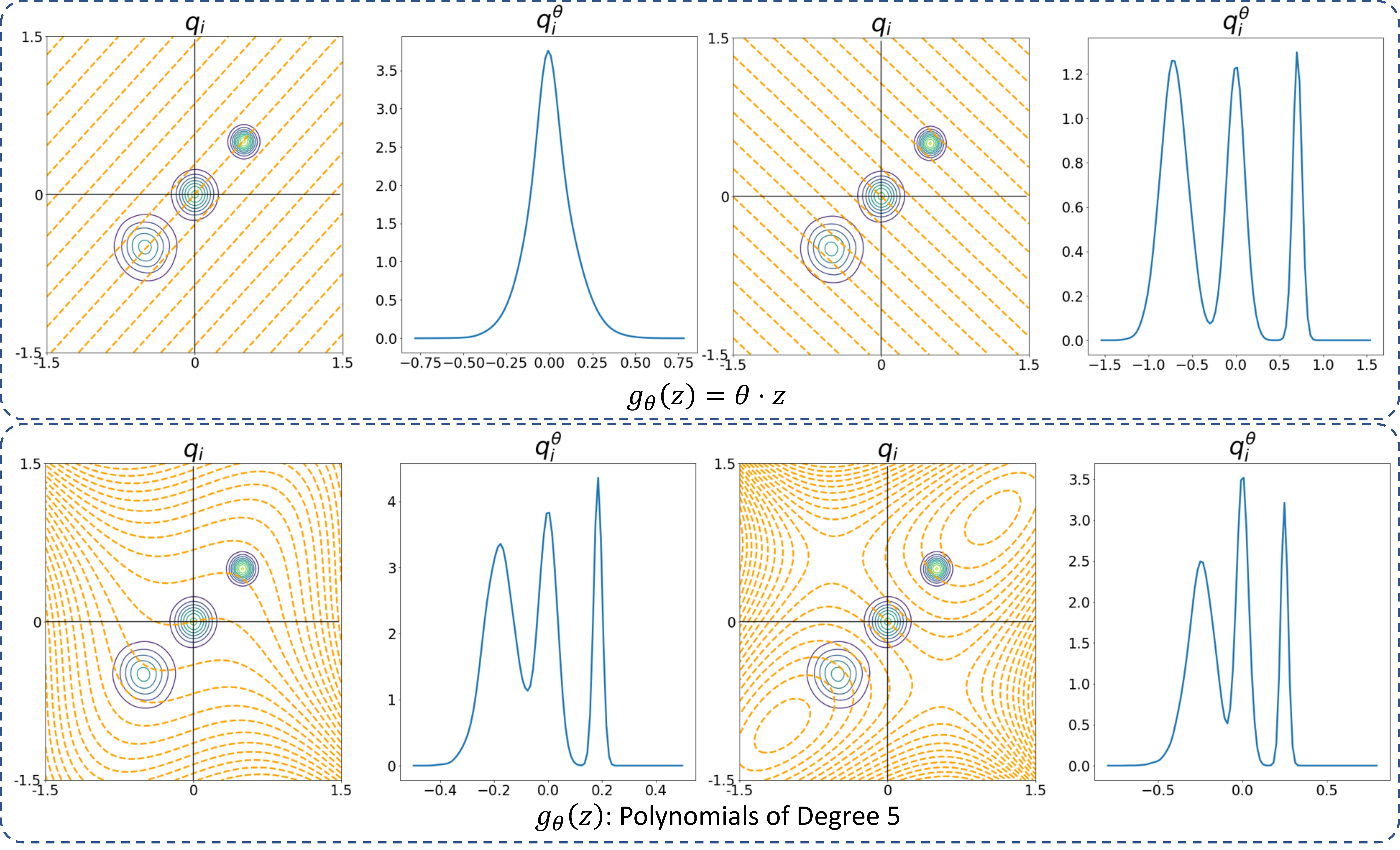}
    \vspace{-.1in}
    \caption{Depiction of random slices of distribution $q_i$ following Eq. \eqref{eq:sliced_measure}, for $g_\theta(z)=\theta\cdot z$ (top row) and for $g_\theta(z)$ being a polynomial of degree 5 (bottom row). The dotted orange lines demonstrate the iso-hypersurfaces of function $g_\theta$, where the $d$-dimensional distribution $q_i$ is integrated over to obtain the one-dimensional density $q_i^\theta$.}
    \label{fig:slicing}
\end{figure}

\subsection{Generalized Sliced-Wasserstein Distances}

Let $g_\theta: \mathbb{R}^d \rightarrow \mathbb{R}$ be a parametric function with parameters $\theta\in\Omega_\theta\subseteq\mathbb{R}^{d_\theta}$, satisfying the regularity conditions in both inputs and parameters as presented in \cite{kolouri2019generalized}. Then a generalized slice of probability measure $\mu_i$ with respect to $g_\theta$ is the one-dimensional probability measure $g_{\theta\#}\mu_i$, which has the following density for all $t\in\mathbb{R}$,
\begin{align}\label{eq:sliced_measure}
    q^\theta_i(t)= \int_{\mathcal{Z}} q_i(z)\delta(t-g_\theta(z))dz,
\end{align}
where $\delta(\cdot)$ denotes the Dirac function on $\mathbb{R}$ (see Figure \ref{fig:slicing}). Having~\eqref{eq:sliced_measure}, the generalized sliced-Wasserstein distance is defined as
\begin{align}
    \mathcal{GSW}_p(\mu_i,\mu_j)=\left(\int_{\Omega_\theta} \mathcal{W}^p_p(g_{\theta\#}\mu_i,g_{\theta\#}\mu_j)d\theta \right)^{\frac{1}{p}}.
    \label{eq:GSW}
\end{align}
Note that for $g_\theta(z)=\theta\cdot z$ and $\Omega_\theta=\mathbb{S}^{d-1}$, where $\mathbb{S}^{d-1}$ denotes the unit $d$-dimensional hypersphere, the generalized sliced-Wasserstein distance is equivalent to the sliced-Wasserstein distance. Equation \eqref{eq:GSW} is the expected value of the Wasserstein distances between slices of distributions $\mu_i$ and $\mu_j$. It has been shown in~\cite{deshpande2019max,kolouri2019generalized} that the expected value in \eqref{eq:GSW} could be substituted with a maximum, i.e., 
\begin{align}
    \text{max-}\mathcal{GSW}_p(\mu_i,\mu_j)= \max_{\theta\in\Omega_\theta} \mathcal{W}_p(g_{\theta\#}\mu_i,g_{\theta\#}\mu_j),
\end{align}
and that max-GSW remains to be a proper statistical metric. Other notable extensions of the GSW distance include the subspace-robust Wasserstein distance \cite{paty2019subspace}, which generalizes the notion of slicing to a projection onto subspaces, and the distributional sliced-Wasserstein distance \cite{nguyen2020distributional} that proposes to replace the expectation with respect to the uniform distribution on $\Omega_\theta$ with a non-uniform distribution.   

From an algorithmic point of view, the expectation in~\eqref{eq:GSW} is approximated using Monte-Carlo integration, which results in an average of a set of $p$-Wasserstein distances between random slices of $d$-dimensional measures. In practice, however, GSW distances only output a good Monte-Carlo approximation using a large number of slices, while max-GSW distances achieve similar results with only a single slice, although at the cost of an optimization over $\theta$. 

\begin{figure*}[t!]
\centering
\includegraphics[trim=.3in 2.5in .3in 2.5in, clip, width=\linewidth]{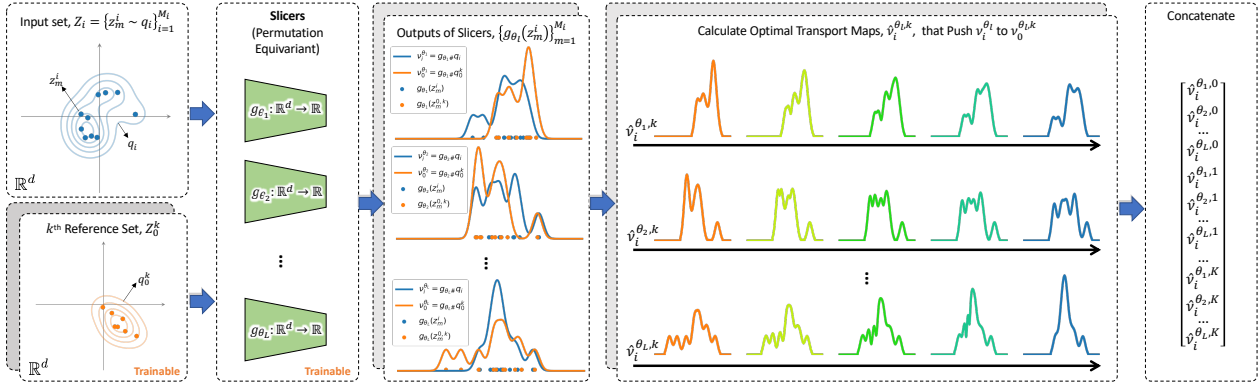}
\caption{An overview of the proposed GSWE framework. Each $d$-dimensional element in a given input set $Z_i$, as well as each reference set $Z_0^k, k\in\{1,\dots,K\}$ is passed through multiple slicers $\{g_{\theta_l}\}_{l=1}^L$. For each slicer, we then perform interpolation on the slicer outputs and derive the optimal transport maps that push the slicer output distributions of a given set to the slicer output distributions of each of the reference sets via~\eqref{eq:monge},~\eqref{eq:embedding_sorting}. The resultant transport maps are then concatenated across all slices and reference sets to derive the final set embeddings. Note that the slicer operations on the input set elements are permutation-equivariant. Moreover, in practice, the slicers may be implemented using multi-layer perceptrons (MLPs), which share the parameters of all layers except for the last layer. In other words, the set of $L$ slicers can be viewed as a unified MLP, represented by a mapping $g: \mathbb{R}^d\rightarrow \mathbb{R}^L$.}
\label{fig:framework}
\vspace{.1in}
\end{figure*}

\section{Generalized Sliced-Wasserstein Embedding}
In this paper, similar to the work of \citet{kusner2015word}, we view the elements of a set as samples from an underlying probability distribution. We then measure the dissimilarity between two sets as the Generalized Sliced-Wasserstein (GSW) distance. Calculating the pairwise distances and leveraging kernel methods, however, would require a quadratic number of distance calculations during training, and a linear number of distance calculations during evaluation (in number of training sets). Instead, here we propose a Euclidean embedding for the GSW distance. We show that this embedding could be thought as a pooling operator, and can be easily incorporated as a layer in a deep neural architecture. Below we describe our proposed embedding. 

We are interested in finding a Euclidean embedding for probability measures, such that the weighted $\ell_p$ distance between two embedded measures is equivalent to the GSW distance between them. Consider a set of probability measures $\{\mu_i\}_{i=1}^N$ with densities $\{q_i\}_{i=1}^N$, 
%We first start with calculating such an embedding for the GSW distance. For
and for simplicity of notation, let $\nu_i^\theta:= g_{\theta\#} \mu_i$ denote the slice of measure $\mu_i$ with respect to $g_{\theta}$. Also, let $\mu_0$ denote a reference measure, with $\nu_0^\theta$ representing its corresponding slice. %\textcolor{red}{Requires explanation}
Then, it is straightforward to show that the optimal transport map (i.e., Monge map) between $\nu_i^\theta$ and $\nu_0^\theta$ can be written as:% (\textcolor{blue}{see supplementary material}): 
\begin{align}
    f^\theta_i = F_{\nu_i^\theta}^{-1}\circ F_{\nu_0^\theta},
    \label{eq:monge}
\end{align}
where as mentioned before, $F_{\nu_i^\theta}^{-1}$ and $F_{\nu_0^\theta}^{-1}$ respectively denote the quantile functions of $\nu_i^\theta$ and $\nu_0^\theta$. Now, letting $id$ denote the identity function, we can write the so-called cumulative distribution transform (CDT) \cite{park2018cumulative} of $\nu_i^\theta$ as
\begin{align}
\hat{\nu}_i^\theta \coloneqq f_i^\theta - id,
\label{eq:cdt}
\end{align}
which, for $p\geq 1$ and for a fixed $\theta$, satisfies the following conditions:
\begin{enumerate}
    \item [C1:]  The weighted $p$-norm of $\hat{\nu}_i^\theta$ equals the  $p$-Wasserstein distance between $\nu_i^\theta$ and $\nu_0^\theta$, i.e.,
    $$\|\hat{\nu}_i^\theta\|_{\nu_0^\theta,p}=\mathcal{W}_p(\nu_i^\theta,\nu_0^\theta),$$
    hence implying that $\|\hat{\nu}_0^\theta\|_{\nu_0^\theta,p}=0$.
    \item [C2:] the weighted $\ell_p$ distance between $\hat{\nu}_i^\theta$ and $\hat{\nu}_j^\theta$ equals the $p$-Wasserstein distance between $\nu_i^\theta$ and $\nu_j^\theta$, i.e.,
    $$\|\hat{\nu}_i^\theta-\hat{\nu}_j^\theta\|_{\nu_0^\theta,p}=\mathcal{W}_p(\nu_i^\theta,\nu_j^\theta).$$
\end{enumerate}
Please refer to the supplementary materials, for a proof of conditions C1 and C2. Finally, the GSW distance between two measures, $\mu_i$ and $\mu_j$, can be obtained as
\begin{align}
&\mathcal{GSW}_p(\mu_i,\mu_j)\nonumber\\
&\quad=\left(\int_{\Omega_\theta}\|\hat{\nu}_i^\theta-\hat{\nu}_j^\theta\|_{\nu_0^\theta, p}^p d\theta\right)^{\frac{1}{p}}\nonumber\\ 
&\quad=\left(\int_{\Omega_\theta}
\left(\int_{\mathbb{R}}\|\hat{\nu}_i^\theta(t)-\hat{\nu}_j^\theta(t)\|_p^p d\nu_0^\theta(t) \right) d\theta\right)^{\frac{1}{p}}.\label{eq:embedding}
\end{align}
Based on \eqref{eq:embedding}, for probability measure $\mu_i$, the mapping to the embedding space is obtained via $\phi(\mu_i)\coloneqq\{\hat{\nu}_i^\theta \}_{\theta\in\Omega_\theta}$. %Note that the results presented in this section hold independent of the choice of $\mu_0$.

\subsection{Empirical Embedding} 
In practice, one often has access only to a finite number of samples from the distributions. % Here we show the empirical version of the GSW embedding. Let 
Specifically, let $Z_i=\{z_m^i \sim q_i\}_{m=1}^{M_i}$ denote the set of $M_i$ samples from the $i$\textsuperscript{th} distribution, and similarly let $Z_0=\{z_m^0 \sim q_0\}_{m=1}^{M}$ denote the set of $M$ samples from the reference distribution. Let $\Theta_L=\{\theta_l\sim \mathcal{U}_{\Omega_\theta}\}_{l=1}^L$ denote a set of $L$ parameter sets sampled uniformly at random from $\Omega_\theta$. Then, the empirical distribution of the $l$\textsuperscript{th} slice of $q_i$ can be written as
\begin{align}
\tilde{q}_i^{\theta_l}=\frac{1}{M_i}\sum_{m=1}^{M_i} \delta(t-g_{\theta_l}(z_m^i)).
\end{align}
In the cases where $M=M_i$, the optimal transport map $f_i^{\theta_l}$ in~\eqref{eq:monge} is obtained by sorting $Z^{\theta_l}_i\coloneqq\{g_{\theta_l}(z_m^i)\}_{m=1}^{M}$, and the embedding can be written as
\begin{align}\label{eq:embedding_sorting}
[\hat{\nu}_i^{\theta_l}]_m = g_{\theta_l}(z_{\pi_i(m)}^i)-g_{\theta_l}(z_{\pi_0(m)}^0),
\end{align}
where $\pi_i(m)$ denotes the permutation obtained by sorting $Z^{\theta_l}_i$. In the cases where $M\neq M_i$, the transport map can be obtained via numerical interpolation using $\eqref{eq:monge}$.

Having the embedding per slice, we can then calculate the empirical GSW distance as
\begin{eqnarray*}
\mathcal{GSW}_p(\mu_i,\mu_j)&=&\|\phi(\mu_i)-\phi(\mu_j)\|_{p,\mu_0}\\ &\approx& \left(\frac{1}{ML} \sum_{l=1}^L \|\hat{\nu}_i^{\theta_l}-\hat{\nu}_j^{\theta_l} \|_p^p\right)^\frac{1}{p}.
\end{eqnarray*}

Note that the aforementioned embedding procedure can be generalized to an arbitrary number of reference sets. Figure~\ref{fig:framework} illustrates an overview of the empirical embedding framework with $K$ reference sets, where the embedding vectors with respect to all reference sets are concatenated to derive the final set embedding. Moreover, to reduce the number of trainable parameters across the $L$ different slices, one can use parameter sharing for a subset of the slice parameter sets $\{\theta_l\}_{l=1}^{L}$. In particular, if each slice is represented by the set of parameters of a neural network, such as a multi-layer perceptron (MLP), with $d$-dimensional inputs and scalar outputs, then the $L$ different neural networks corresponding to the $L$ slices may share their parameters in all layers except for the last layer. This is equivalent to combining all the slicers into an aggregate neural network with $d$-dimensional inputs and $L$-dimensional outputs.

Given the high-dimensional nature of the problems of interest in machine learning, one often requires a large number of random samples, $L$, to obtain a good approximation of the GSW distance. This is related to the projection complexity of the sliced distances \cite{deshpande2019max}. To avoid the poor scaling of Monte-Carlo approximation with respect to the number of slices, we devise a unique approach that ties GSW embedding to metric learning. First, we note that ideas like max-GSW \cite{kolouri2019generalized, deshpande2019max} or subspace-robust Wasserstein distance \cite{paty2019subspace} would not be practical in this setting, where the slicing parameters, $\Theta_L$, are fixed for all probability measures and not chosen separately for each probability measure $\mu_i$. Next we propose a solution to this problem.

\subsection{Optimal Slices for a Set of Distributions}
\label{sec:ssl}
Given samples from our training probability distributions, i.e., $\{Z_i\}_{n=1}^N$, and samples from a reference measure, $Z_0$, we seek an optimal set of $L$ slices $\Theta^*_L$ that could be learned from the data. The optimization on $\Theta^*_L$ ties the GSWE framework to the field of metric learning, allowing us to find slices or, in other words, an \emph{embedding} with a specific statistical characterization. In the following, we propose two different approaches to finding the optimal slices, which are both rooted in the core idea of contrastive learning, commonly used as a self-supervisory signal. With a slight abuse of notation, for any given distribution $Z_i$ and set of slices $\Theta_L$, we denote the corresponding embedding of $Z_i$ by $\nu_i^{\Theta_L}$.
\begin{itemize}[leftmargin=*]
% \item Find slices that maximize the pairwise GSW distances between all pairs of probability measures in the set, i.e.,
% $$ \operatorname{argmax}_{\Theta_L\in\Omega_\theta} \frac{1}{ML}\sum_i\sum_{j\neq i}\sum_{\theta\in\Theta_L} \|\hat{\nu}_i^{\theta}-\hat{\nu}_j^{\theta} \|_p^p + \lambda \rho(\Theta_L)$$
% where $\rho(\cdot)$ is a regularizer on the parameters, and $\lambda$ is the regularization coefficient. Finding such slices is equivalent to finding an embedding in which every distribution is as far away from other distributions as possible. This idea was used in \cite{wu2018unsupervised} for self-supervised learning.
%    \item For scenarios in which we can assign a set of positive and negative probability measures to $\mu_i$, we leverage the following contrastive loss:
%    $$
%    \operatorname{argmax}_{\Theta_L\in\Omega_\theta}\sum_{\theta\in\Omega_\theta}\sum_i %\frac{1}{P(i)}\sum_{p\in P(i)} \frac{exp(-)}{xxx}
%    $$
\item \textbf{SimCLR.} In scenarios where there exists some a priori notion of similarity and dissimilarity between sets, we leverage the following contrastive loss~\cite{chen2020simple} to find the optimal slices, where for each batch of $\mathcal{B}$ of $|\mathcal{B}|$ samples, the optimization problem can be written as
\begin{align}\label{eq:simclr_main}
\min_{\Theta_L\in\Omega_\theta^L}\frac{1}{2|\mathcal{B}|} \sum_{i\in\mathcal{B}} \left(\ell_i^{\Theta_L} + \overline{\ell}_i^{\Theta_L} \right),
\end{align}
For each sample $i\in\mathcal{B}$, the two loss terms $\ell_i^{\Theta_L}$ and $\overline{\ell}_i^{\Theta_L}$ in~\eqref{eq:simclr_main} are respectively defined as
\begin{align}
\ell_i^{\Theta_L} & \coloneqq -\log \tfrac{\mathcal{S}(\nu_i^{\Theta_L}, \overline{\nu}_i^{\Theta_L})}{\sum_{j\in\mathcal{B}} \mathcal{S}(\nu_i^{\Theta_L}, \overline{\nu}_j^{\Theta_L}) + \sum_{k\in\mathcal{B}\setminus\{i\}} \mathcal{S}(\nu_i^{\Theta_L}, \nu_k^{\Theta_L})}\label{eq:simclr1} \\
\overline{\ell}_i^{\Theta_L} & \coloneqq -\log \tfrac{\mathcal{S}(\overline{\nu}_i^{\Theta_L}, \nu_i^{\Theta_L})}{\sum_{j\in\mathcal{B}} \mathcal{S}(\overline{\nu}_i^{\Theta_L}, \nu_j^{\Theta_L}) + \sum_{k\in\mathcal{B}\setminus\{i\}} \mathcal{S}(\overline{\nu}_i^{\Theta_L}, \overline{\nu}_k^{\Theta_L})},\label{eq:simclr2}
\end{align}
where for a given temperature hyperparameter $\tau$, we define
\begin{align}
\mathcal{S}(x,y)\coloneqq \exp(x^T y / \tau).
\end{align}
In~\eqref{eq:simclr1}-\eqref{eq:simclr2}, for each sample $i\in\mathcal{B}$, $\overline{\nu}_i^{\Theta_L}$ denotes the embedding of some $\overline{Z}_i$ similar to $Z_i$ (usually generated from $Z_i$ via an augmentation procedure), while for any $j\in\mathcal{B}\setminus\{i\}$, both $Z_j$ and $\overline{Z}_j$ are assumed to be dissimilar to both $Z_i$ and $\overline{Z}_i$.

\item\textbf{SimSiam.} We also examine a more recent self-supervised formulation by~\cite{chen2020exploring}, in which only positive examples factor into the loss. Specifically, for a batch of samples denoted by $\mathcal{B}$, the optimal slices are found by solving the following optimization problem,
\begin{align}
\min_{\Theta_L\in\Omega_\theta^L} \frac{1}{2|\mathcal{B}|}\sum_{i\in\mathcal{B}} \left(\mathcal{D}(\nu_i^{\Theta_L}, \overline{\nu}_i^{\Theta_L}) + \mathcal{D}(\overline{\nu}_i^{\Theta_L} \nu_i^{\Theta_L})\right),
\end{align}
where
\begin{align}
\mathcal{D}(x, y) \coloneqq \|x-\operatorname{stopgrad}(y) \|_p^p,
\end{align}
and as before, $\overline{\nu}_i^{\Theta_L}$ represents the embedding of an augmentation of $\nu_i^{\Theta_L}$. The operator $\operatorname{stopgrad}(\cdot)$ is included to prevent mode collapse in the latent space due to the lack of negative samples in this loss formulation.
\end{itemize}

\begin{figure*}[t]
\centering
\includegraphics[width=\linewidth]{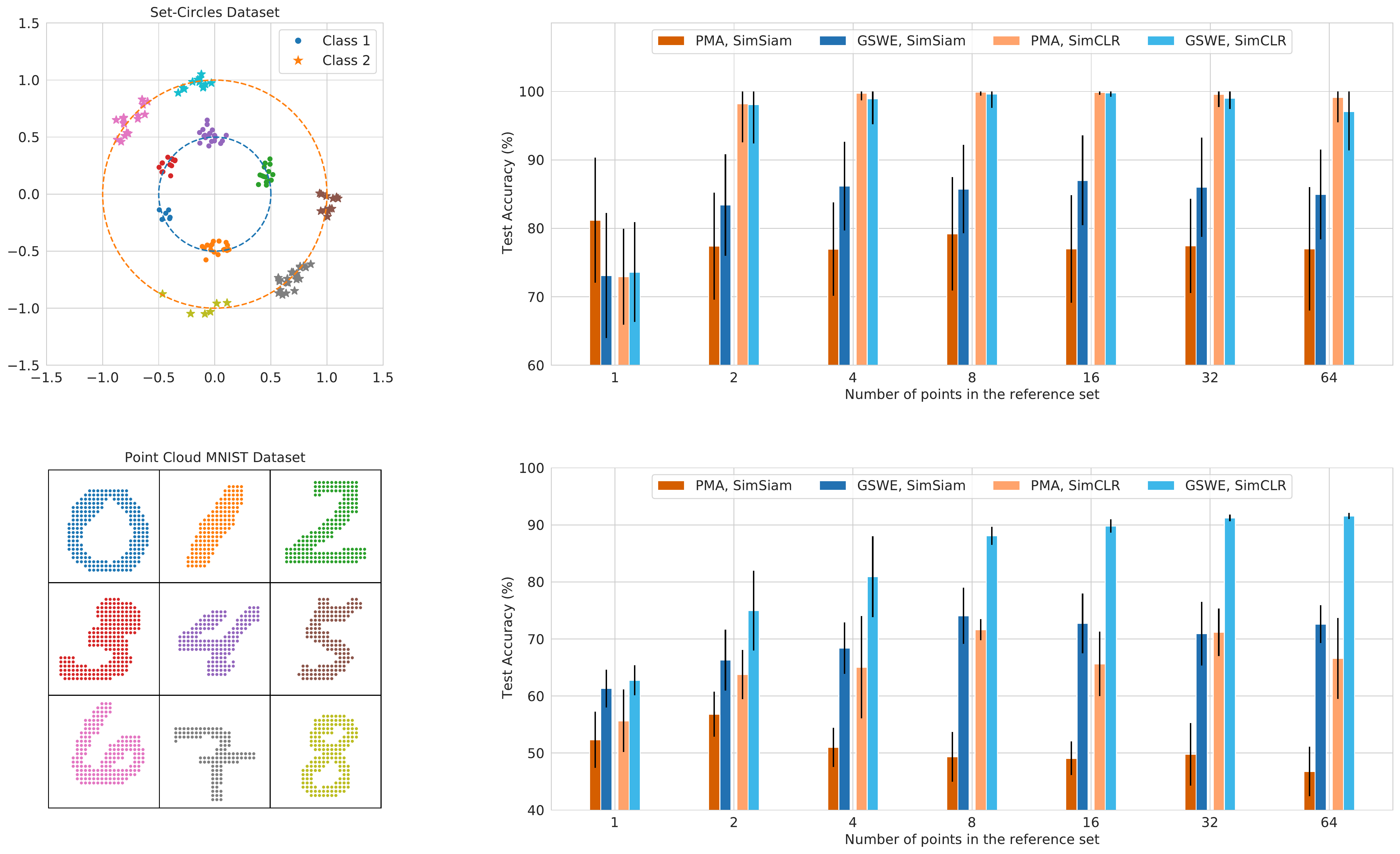}
\caption{Sample sets from the \texttt{Set-Circles} dataset (top left) and the \texttt{Point Cloud MNIST} dataset (bottom left), alongside the nearest neighbor (NN) test accuracies of Pooling with Multi-head Attention (PMA) and our proposed method, GSWE, using different self-supervised loss functions and reference set cardinalities.}
\label{fig:unsupervised_datasets_results}
\end{figure*}

\begin{figure*}[t]
\centering
\includegraphics[width=\textwidth]{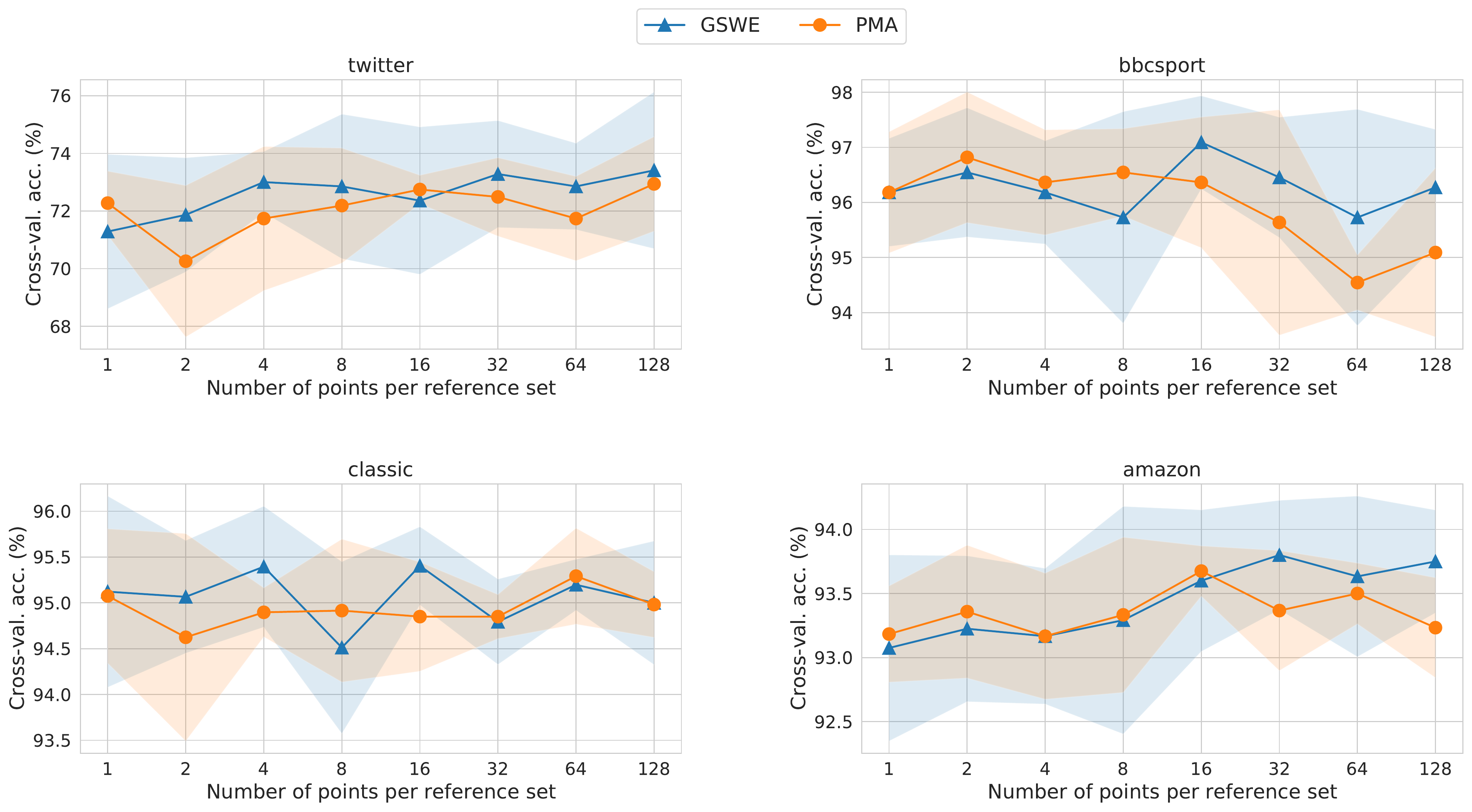}
\caption{The 5-fold cross-validation accuracy of GSWE as compared to PMA in the supervised learning setting on four text categorization datasets using a single reference set with varying number of elements.}
\label{fig:supervised_1refSet}
\end{figure*}

\begin{figure*}[t]
\centering
\includegraphics[width=\textwidth]{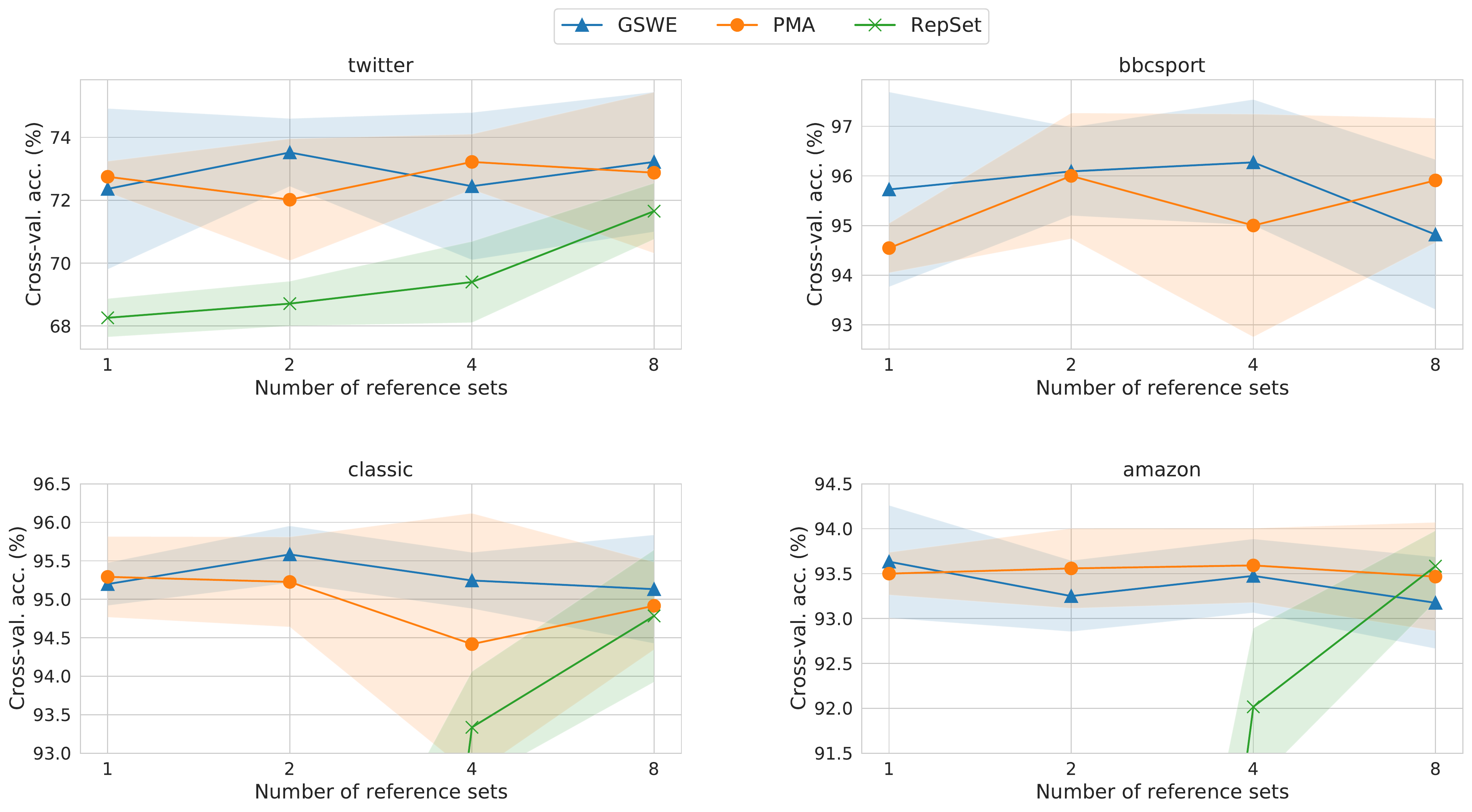}
\caption{The 5-fold cross-validation accuracy of GSWE, PMA, and RepSet in the supervised learning setting on four text categorization datasets using varying number of reference sets with a fixed number of elements per reference set (16 in case of \texttt{twitter}, and 64 for the other three datasets).}
\label{fig:supervised_fixed_point_per_refSet}
\end{figure*}

\section{Experiments}

We evaluate the proposed GSWE algorithm against Set Transformer~\cite{lee2019set} and RepSet \cite{skianis2020rep} baseline methods, for both unsupervised (more precisely self-supervised) and supervised learning on various set-structured datasets. Implementation details of the experiments can be found in the Supplementary Material.

\subsection{Unsupervised Learning}\label{sec:unsupervised}

%Here we focus on unsupervised learning on set-structured data. We leverage the unsupervised learning concepts in self-supervised learning as discussed in Section \ref{sec:ssl}, and apply these ideas to learning from sets. 

We first focus on unsupervised learning of set embeddings using the self-supervised approaches discussed in Section~\ref{sec:ssl}. We consider the following two datasets:
\begin{itemize}[leftmargin=*]
\item \texttt{Set-Circles}: We generate a simple two-dimensional dataset called, Set-Circles (see Figure \ref{fig:unsupervised_datasets_results} (top left)). Each sample is a set with random number of elements generated on a random arc on a circle with additive noise. There are two classes, in the dataset which are identified by the radius of the circle the samples live on. The average norm of the set elements is an ideal feature for discriminating the classes.

\item \texttt{Point Cloud MNIST}: We also consider the two-dimensional point cloud MNIST dataset~\cite{lecun1998gradient}, where each sample consists of a set of points in the xy-plane converted from the pixels of a corresponding 0-9 digit image (see Figure \ref{fig:unsupervised_datasets_results} (bottom left)).
\end{itemize}

We perform self-supervised learning on these datasets using the two losses covered in Section \ref{sec:ssl}, namely SimCLR and SimSiam. After training the networks, we perform nearest neighbor retrieval for the test sets and measure the label agreement between the input and the retrieved set. We emphasize that for the \texttt{Set-Circles} dataset, in order to avoid a trivial solution, all methods use a backbone that maps the set elements from $\mathbb{R}^2$ to $\mathbb{R}$ (Otherwise the problem becomes trivial to solve).

Figure~\ref{fig:unsupervised_datasets_results} shows the performance of GSWE as compared with the Set Transformer architecture (denoted by PMA, referring to the pooling with multi-head attention module) for different cardinalities of the reference set. As the figure shows, on both datasets and using both loss functions, our proposed approach either performs similarly to or outperforms the Set Transformer method for reference sets with greater than a single element. Note that for a single element in the reference set, our proposed GSWE method effectively reduces to global average pooling, while PMA can be viewed as \emph{weighted} global average pooling. That explains the performance gain achieved by PMA for a reference set of size 1 with SimSiam loss on the \texttt{Set-Circles} dataset. However, as soon as an additional element is added to the reference set, GSWE performs significantly better than PMA. Note that given the same backbone, GSWE has far fewer parameters than PMA, due to the absence of multi-head attention in the pooling module in GSWE, which helps explain the superiority of GSWE as compared to PMA in terms of nearest neighbor retrieval accuracy.

\subsection{Supervised Learning}
We also evaluate our proposed method on a set of four text categorization datasets, namely \texttt{twitter}, \texttt{bbcsport}, \texttt{classic}, and \texttt{amazon}.~\cite{skianis2020rep}. In each dataset, each input sample is a document, viewed as a set of elements, where each element corresponds to a 300-dimensional word embedding of a term in the document, and the goal is to classify the documents based on the word embedding sets in a supervised manner.

For these experiments, we report the 5-fold cross-validation accuracy using our proposed GSWE method, as compared to PMA, i.e., Set Transformer, and RepSet. Note that the number of reference sets in our approach and RepSet is analogous to the number of attention heads in PMA. Moreover, the number of the points/elements per reference set is analogous to the number of seeds in PMA. We perform a comparative study to demonstrate the performance of each method as a function of these parameters. More details on the experiments can be found in the Supplementary Material.

Figure~\ref{fig:supervised_1refSet} shows the 5-fold cross-validation accuracy achieved by GSWE and PMA for different numbers of elements in a single reference set. As RepSet is designed only for multiple reference/hidden sets, we omit its performance results from this figure. As the figure demonstrates, across all datasets, GSWE performs on par with PMA, while exhibiting superior peak accuracies as compared to PMA.% for a single element per reference set, PMA outperforms GSWE across all four datasets, which is consistent with our observations in Section~\ref{sec:unsupervised}. However, GSWE performs better for larger reference sets and performs similarly to PMA on most datasets, except for \texttt{twitter}, on which it outperforms PMA.

Moreover, Figure~\ref{fig:supervised_fixed_point_per_refSet} shows the performance of GSWE, PMA, and RepSet for different numbers of reference sets, where the cardinality of each reference set is fixed at 16 for \texttt{twitter} (due to smaller set cardinalities) and 64 for the remaining datasets. As the figure demonstrates, while both GSWE and PMA demonstrate a robust performance with respect to the number of reference sets, with GSWE again having the edge over PMA in terms of peak accuracy, RepSet critically depends on having a higher number of reference sets, and having few reference sets dramatically reduces its achievable accuracy, which is why we have omitted its performance results for the cases where it was far inferior than that of GSWE and PMA.

\section{Conclusion}
We introduced a novel method for learning representations from set-structured data via generalized sliced Wasserstein (GSW) distances. Our method treats the elements of each input set as samples from a distribution, and derives an embedding for the entire set based on the GSW distance between the representations of the set elements (derived through a permutation-equivariant backbone) and one or multiple reference set(s), whose elements are learned in an end-to-end fashion. We showed that our method derives an exact Euclidean embedding which is geometrically-interpretable for set-structured data. Moreover, we demonstrated, through experimental results, that our set embedding approach provides state-of-the-art performance on a variety of supervised and unsupervised set classification tasks, in part due to a reduced number of parameters as opposed to attention-based pooling methods.

\section*{Acknowledgement}
This material is based upon work supported by the United States Air Force under Contract No. FA8750‐19‐C‐0098. Any opinions, findings, and conclusions or recommendations expressed in this material are those of the author(s) and do not necessarily reflect the views of the United States Air Force and DARPA.

% \balance
\bibliography{gswe}
\bibliographystyle{icml2021.bst}

\begin{appendices}
\section{Implementation Details}

\subsection{Unsupervised Experiments}
For the \texttt{set-circles} dataset, we use a multi-layer perceptron (MLP) backbone with $2$ hidden layers, each of size $64$, rectified linear unit (ReLU) non-linearity, and output size of $1$ (corresponding to a single slice, as adding more outputs makes the classification problem trivial). To create augmentations, we rotate the elements of each set by a certain angle, uniformly selected at random from the interval $[0,2\pi)$. Training is conducted for $50$ epochs, using a batch size of $32$ and Adam optimizer with a learning rate of $10^{-4}$. The training process is repeated $100$ times, each with a different random seed.

For the \texttt{point} \texttt{cloud} \texttt{MNIST} dataset, inspired by~\cite{kosiorek2020conditional}, we use an attention-based backbone using the set attention block (SAB) module introduced in~\cite{lee2019set} for both the GSWE and PMA pooling methods, where consecutive layers share their parameters. In particular, the $2$-dimensional input feature is first projected into a $256$-dimensional space through a linear mapping. It then undergoes a $256$-dimensional SAB layer with $4$ attention heads $3$ consecutive times, and the output is then projected to a $16$-dimensional output using a final linear mapping. To create augmentations, we perturb the $(x,y)$ coordinates of each element using Gaussian noise with zero mean and unit variance. Training is conducted for $25$ epochs, using a batch size of $32$ and Adam optimizer with a learning rate of $10^{-3}$. The training process is repeated $10$ times, each with a different random seed.

For both datasets and both pooling mechanisms, we set the temperature hyperparamter for the contrastive loss to $\tau=0.1$. Moreover, once training is completed, we freeze the set embeddings, evaluate the 1-nearest neighbor (1-NN) accuracy of the test samples (using neighbors from the training samples), and report the mean and standard deviation of the resulting accuracies across the runs with different random seeds.

\aboverulesep=0ex
\belowrulesep=0ex
\renewcommand{\arraystretch}{1.2}
\begin{table*}[t]
\scriptsize
\centering
\noindent\makebox[\textwidth]{
\rowcolors{2}{Gray}{}
\setlength\tabcolsep{6pt}
\begin{tabular}{c|ccccc}
\cmidrule[1.5pt]{1-6}
Experiment type & Dataset & Number of training sets & Mean training set size & Number of features per element & Number of classes\\ 
\cline{1-6}
& \texttt{Set-Circles} & \phantom{60,}400 & \phantom{6}14.6 & 2 & 2 \\
\multirow{-2}{*}{\cellcolor{white}Unsupervised} & \texttt{Point} \texttt{Cloud} \texttt{MNIST} & 60,000 & 149.9 & 2 & 10 \\
\hhline{|*6{-}|}
\cellcolor{white}& \texttt{twitter} & \phantom{6}2,176 & \phantom{60}9.9 & 300 & 3 \\
& \texttt{bbcsport} & \phantom{60,}517 & 117.4 & 300 & 5 \\
\cellcolor{white} & \texttt{classic} & \phantom{6}4,965 & \phantom{6}38.6 & 300 & 4 \\
\multirow{-4}{*}{\cellcolor{white}Supervised} & \texttt{amazon} & \phantom{6}5,600 & \phantom{6}45.0 & 300 & 4 \\
\cmidrule[1.5pt]{1-6}
\end{tabular}}
\caption{Statistics of the datasets used in the unsupervised and supervised experiments.}
\label{tab:datasets}
\end{table*}

\subsection{Supervised Experiments}
For all datasets, we consider an attention-based backbone using the set attention block (SAB) module~\cite{lee2019set} for the GSWE and PMA pooling methods. In particular, we use a backbone with two $128$-dimensional hidden layers and one $16$-dimensional output layer. The hidden layers use $4$ attention heads, while the last layer uses a single attention head. After the pooling modules, we use a classifier with a single $128$-dimensional hidden layer and rectified linear unit (ReLU) non-linearity. For evaluating the RepSet baseline, we use the same end-to-end architecture as in~\cite{skianis2020rep}, including the classifier and the network-flow-based backbone. All algorithms are trained for $50$ epochs, using a batch size of $64$ and Adam optimizer with a learning rate of $10^{-3}$, and the 5-fold cross-validation accuracy is reported.

\section{Dataset Statistics}
Table~\ref{tab:datasets} shows the statistics of the two point cloud datasets and four text categorization datasets that we used for the unsupervised and supervised experiments, respectively.

\section{Proof of C1 and C2}

First we show that the reference is mapped to the origin, $\hat{\nu}_0^\theta=\bf{0}$.  
\begin{align*}
    \hat{\nu}_0^\theta&= F^{-1}_{\nu_0^\theta}\circ F_{\nu_0^\theta}-id\\
    &= id-id = \bf{0},
\end{align*}
where we used definition \eqref{eq:cdt}, and Equation \eqref{eq:monge}. Now we prove C2.
{\small
\begin{align*}
\|\hat{\nu}_i^\theta-\hat{\nu}_j^\theta\|_{\nu_0^\theta,p}&=\|f_i^\theta-f_j^\theta\|_{\nu_0^\theta,p}\\
&= \left(\int_\mathbb{R} \|f_i^\theta(t)-f_j^\theta(t)\|^p d\nu_0^\theta(t) \right)^{\frac{1}{p}}\\
&= \left(\int_\mathbb{R} \|F^{-1}_{\nu_i^\theta}(F_{\nu_0^\theta}(t))-F^{-1}_{\nu_j^\theta}(F_{\nu_0^\theta}(t))\|^p d\nu_0^\theta(t) \right)^{\frac{1}{p}}\\
&= \left(\int_0^1 \|F^{-1}_{\nu_i^\theta}(u)-F^{-1}_{\nu_j^\theta}(u)\|^p du \right)^{\frac{1}{p}}\\
&=\mathcal{W}_p(\nu_i^\theta,\nu_j^\theta).
\end{align*}}
Finally, given C2 and the fact that $\hat{\nu}_0^\theta=\bf{0}$, we obtain C1: 
\begin{align*}
\|\hat{\nu}_i^\theta\|_{\nu_0^\theta,p}=\|\hat{\nu}_i^\theta-\hat{\nu}_0^\theta\|_{\nu_0^\theta,p}=\mathcal{W}_p(\nu_i^\theta,\nu_0^\theta).
\end{align*}
\end{appendices}

%%%%%%%%%%%%%%%%%%%%%%%%%%%%%%%%%%%%%%%%%%%%%%%%%%%%%%%%%%%%%%%%%%%%%%%%%%%%%%%
%%%%%%%%%%%%%%%%%%%%%%%%%%%%%%%%%%%%%%%%%%%%%%%%%%%%%%%%%%%%%%%%%%%%%%%%%%%%%%%
% DELETE THIS PART. DO NOT PLACE CONTENT AFTER THE REFERENCES!
%%%%%%%%%%%%%%%%%%%%%%%%%%%%%%%%%%%%%%%%%%%%%%%%%%%%%%%%%%%%%%%%%%%%%%%%%%%%%%%
%%%%%%%%%%%%%%%%%%%%%%%%%%%%%%%%%%%%%%%%%%%%%%%%%%%%%%%%%%%%%%%%%%%%%%%%%%%%%%%
% \appendix
% \section{Do \emph{not} have an appendix here}

% \textbf{\emph{Do not put content after the references.}}
% %
% Put anything that you might normally include after the references in a separate
% supplementary file.

% We recommend that you build supplementary material in a separate document.
% If you must create one PDF and cut it up, please be careful to use a tool that
% doesn't alter the margins, and that doesn't aggressively rewrite the PDF file.
% pdftk usually works fine. 

% \textbf{Please do not use Apple's preview to cut off supplementary material.} In
% previous years it has altered margins, and created headaches at the camera-ready
% stage. 
%%%%%%%%%%%%%%%%%%%%%%%%%%%%%%%%%%%%%%%%%%%%%%%%%%%%%%%%%%%%%%%%%%%%%%%%%%%%%%%
%%%%%%%%%%%%%%%%%%%%%%%%%%%%%%%%%%%%%%%%%%%%%%%%%%%%%%%%%%%%%%%%%%%%%%%%%%%%%%%

\end{document}